\documentclass[letterpaper, 10 pt, conference]{ieeeconf}  

\IEEEoverridecommandlockouts                              

\usepackage[hidelinks]{hyperref}
\usepackage{todonotes}

\newcommand{\etal}{\emph{et al.}}
\newcommand{\algname}{FM-Loc}

\title{\LARGE \bf
\algname: Using Foundation Models \\ for Improved Vision-based Localization
}

\author{Reihaneh Mirjalili$^{1}$, Michael Krawez$^{1}$ and Wolfram Burgard$^{1}$
\thanks{$^{1}$All authors are with the Department of Engineering, University of
Technology Nuremberg, Germany.
        }%
}
\usepackage{graphicx}
\usepackage[export]{adjustbox}
\usepackage{amsmath}

\usepackage[T1]{fontenc}
\usepackage{array}
\usepackage{booktabs}

\usepackage{pbox}
\begin{document}
\maketitle
\thispagestyle{empty}
\pagestyle{empty}
\begin{abstract}
Visual place recognition is essential for vision-based robot localization and SLAM. Despite the tremendous progress made in recent years, place recognition in changing environments remains challenging. A promising approach to cope with appearance variations is to leverage high-level semantic features like objects or place categories. In this paper, we propose \algname{} which is a novel image-based localization approach based on Foundation Models that uses the Large Language Model GPT-3 in combination with the Visual-Language Model CLIP to construct a semantic image descriptor that is robust to severe changes in scene geometry and camera viewpoint. We deploy CLIP to detect objects in an image, GPT-3 to suggest potential room labels based on the detected objects, and CLIP again to propose the most likely location label. The object labels and the scene label constitute an image descriptor that we use to calculate a similarity score between the query and database images. We validate our approach on real-world data that exhibit significant changes in camera viewpoints and object placement between the database and query trajectories. The experimental results demonstrate that our method is applicable to a wide range of indoor scenarios without the need for training or fine-tuning.



\end{abstract}


\section{INTRODUCTION}
Robust place recognition is of utmost importance for robot navigation as it supports highly relevant tasks including localization and SLAM. To efficiently perform navigation tasks and to build consistent maps of their environment, robots need the ability to determine their location given a map or a set of previously recorded observations. In vision-based place recognition, one relies on camera images and typically matches a query image to a set of reference or previously recorded images, which is akin to the problem of image retrieval. A popular approach to vision-based place recognition is to use local image features, While such feature-based methods provide good results in many applications, they typically are less robust when the scene appearance undergoes substantial changes.  In indoor environments, such changes can come either from lighting variations or from the addition, removal, or rearrangement of objects within the scene. Further problems for feature-based methods are caused by substantial viewpoint changes between the individual observations. A recent trend to tackle vision-based place recognition under such dynamic changes is to incorporate high-level semantic information, e.g., through the introduction of objects~\cite{bowman2017probabilistic, yu2018ds, gomez2020object}. 

\begin{figure}[t]
\vspace{1ex}
\centering
\includegraphics[clip,trim=2cm 0cm 1cm 0cm,scale=0.28]{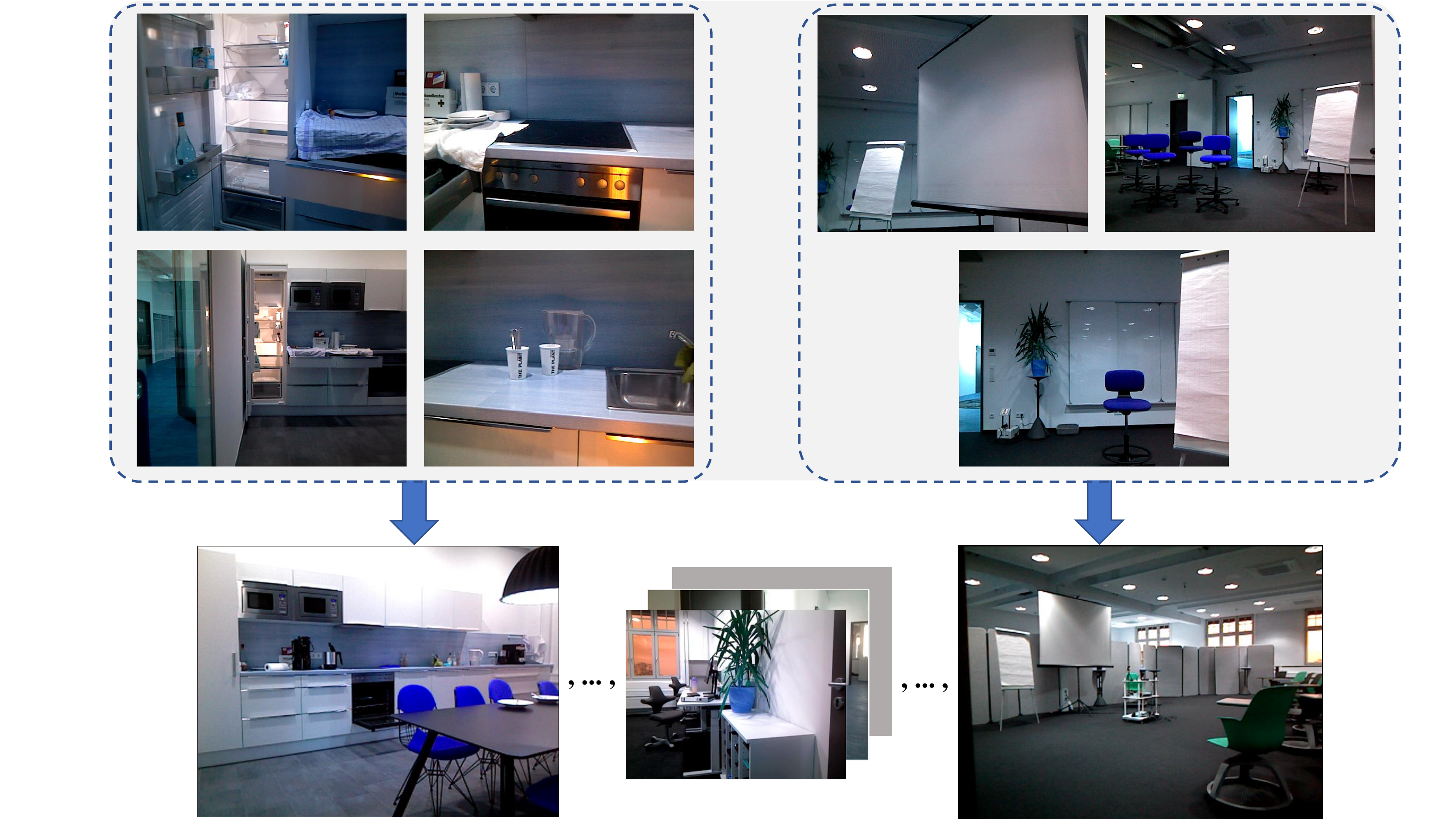}  
\caption{Our localization method matches query images (top) to a set of reference images (bottom). Despite considerable differences in viewpoint and object placement between reference and query image sets, our approach correctly recognizes the locations.}
\label{fig:covergirl}
\end{figure}

In this paper, we propose \algname{} which is a novel method based on Foundation Models for vision-based localization in changing indoor environments. We leverage Large Language Models (LLMs) and Visual-Language Models (VLMs) to derive a semantic image descriptor that is robust to extreme scene rearrangements and viewpoint changes. Our method uses the VLM CLIP to detect objects in query and reference images and the LLM GPT-3 to classify the location based on the extracted object labels. It uses the object and location labels to form a semantic image descriptor that is robust even to extreme scene rearrangements and viewpoint changes, as shown in \autoref{fig:covergirl}. 

Typically, LLMs like BERT~\cite{devlin2018bert} or GPT-3~\cite{brown2020language} and VLMs like CLIP~\cite{CLIP} are referred to as foundation models~\cite{bommasani2021opportunities} since their scale and the generality of the training data enables downstream applications to a wide field of tasks. In conjunction, LLMs and VLMs allow for the grounding of natural language commands or descriptions in the real world, which is an important precondition for many robotics tasks including manipulation and human-robot interaction. While foundation models were already applied successfully in the field of robotics to perform navigation, object detection, and manipulation tasks, their utilization for SLAM and localization has received less attention. In the  localization method proposed in this paper, we employ foundation models to ground object descriptions in images and to classify the depicted locations.

In particular, we deploy CLIP to match an image and a list of object labels and retrieve the objects with the highest matching score. Using the object labels as the sole image descriptor has several caveats. Static and unique objects are good global landmarks, however, other objects might be specific to a room but not static, and yet other objects do not provide any useful information about the location. For example, in an office building chairs can be found at working places, in meeting rooms, or in a kitchen. A set of dynamic objects like cups, plates, and cutlery can indicate a specific room, e.g., a kitchen, but not be present in the same configuration in query and reference images. To robustify the descriptor, we aggregate over the object labels by prompting GPT-3 to provide likely room types that contain these objects. We again use CLIP to ground this candidate list in the image and to select the final room label. The room label and the labels of detected landmark objects then constitute the image descriptor.  

The fact that our approach relies on foundation models makes it applicable to a wide range of environments without the need to re-train or fine-tune the underlying models. Further, LLMs and VLMs permit open-vocabulary lists of object and room labels which can be either drawn from existing taxonomies or defined manually. Thus, labels of descriptive landmarks specific to the target environment can be easily added to the list by the user and our approach does not require any pre-trained classifier to perform the place recognition task. 

In summary, we make the following contributions: 
\begin{enumerate}
    \item We present a novel approach based on foundation models to improve vision-based localization,
    \item we demonstrate that our  approach  can easily be extended to novel landmarks without additional training, and 
    \item we provide an approach to extract the landmarks that are most relevant for the localization task.
\end{enumerate}
 We furthermore present real-world experiments in dynamic environments to  demonstrate these capabilities.


\section{Related Work}
\label{sec:related_work}
Large-scale language models have recently become more common in the field of robotics and recent works demonstrate their potential to simplify robotic problems like navigation or human-robot interaction. The
LM-Nav system~\cite{shah2022robotic} leverages pre-trained models to navigate a robot in a topological map following natural language instructions. The approach first extracts a sequence of landmark labels from the instruction using GPT-3. It then uses CLIP to match  landmark labels and map images to determine likely landmark locations. They employ a graph search algorithm to find a path that visits the nodes containing the requested landmarks. Finally, they employ ViNG~\cite{shah2021ving} to navigate the robot between the graph nodes. 
Huang \etal~\cite{huang2022visual} also tackle the problem of following instructions in robot navigation. They use a grid map annotated with CLIP features extracted from input images. The metric map representation enables the user to include spatial descriptions of the goal location in the instructions.
CLIP-Nav~\cite{dorbala2022clip} uses GPT-3 to break down complex navigation instructions into command keyphrases. It selects the next intermediate goal  using CLIP to ground a keyphrase in the panoramic image of the current location.
Gadre \etal~\cite{gadre2022clip} utilize CLIP for zero-shot, open-vocabulary object detection. Given a target object label, the robot explores its vicinity and generates a top-down map from RGB-D images. The algorithm monitors the CLIP similarity score between the current image and the object label and ends the search if that score is above a certain threshold.

LLMs were further applied in robotic manipulation. Brohan \etal~\cite{brohan2022can} present a scheme for grounding language commands in robot actions. In particular, they use RL to learn an affordance function  that estimates for a set of actions their execution success in a given context. To complete a task at hand, the LLM computes a score for the next best action which is then combined with the affordance function to ensure the  feasibility of the action. Chen \etal~\cite{chen2022open} extend the latter work to scene-level tasks with open vocabulary.

The approaches discussed so far mostly use the notion of objects to ground natural language instructions in the physical world, but do not explicitly address the problem of robot localization.
However, the objects detected by a robot can provide valuable clues about its location. For instance, observing a landmark that is unique in the environment strongly reduces the pose hypothesis space. Yet other, non-unique objects might be specific to a certain room type. Thus, several works use semantic object labels and a language-based classifier to leverage localization or scene recognition. 

Heikel and Espinosa-Leal~\cite{heikel2022indoor} deploy YOLO~\cite{redmon2016you} to retrieve object labels from an input image. They transform these labels into language-based features and employ a random forest classifier to predict the room label. Similarly, Chen \etal~\cite{chen2019scene} obtain object labels from YOLO and use a learned taxonomy to refine the results of a CNN-based scene classifier. Our approach follows the same strategy of deriving the room class from object labels. In contrast to the approaches mentioned above, we employ foundation models for both, object detection and scene classification. This allows us to perform both tasks in a zero-shot fashion and further grants greater flexibility on landmark and environment labels.

Appearance-based localization is closely related to image retrieval and place recognition. Most state-of-the-art approaches in these fields are able to match query images to a reference database if the query and reference images were captured under similar conditions. However, making image retrieval robust to drastic viewpoint and appearance changes is still an ongoing research problem.
A line of work by Vysotska \etal~\cite{vysotska2015efficient, vysotska2016lazy, vysotska2017relocalization} explores image sequence matching under seasonal or day-and-night appearance variations. Tomit{\u{a}} \etal~\cite{tomitua2021convsequential} also exploit sequential information to increase the robustness of place recognition. In single-image retrieval, data-driven approaches dominate the field.
NetVLAD~\cite{arandjelovic2016netvlad} is an end-to-end CNN model that is the de facto standard for place recognition. Patch NetVLAD~\cite{patchnetvlad} improves the results of the basic NetVLAD method by local feature matching. That approach displays significant robustness to viewpoint and appearance variations, although it can still fail under drastic changes. Garg, Suenderhauf, and Milford~\cite{garg2022semantic} show that incorporating semantics can improve opposing view matching in cases of changing scene appearance. Similar in spirit, we propose an approach to combine the Patch NetVLAD score with our language-based semantic descriptor and demonstrate that the matching performance is improved under extreme appearance and viewpoint variations.

\begin{figure*}[t]
\centering
\includegraphics[clip,trim=0 1cm 0 1cm,scale=0.6]{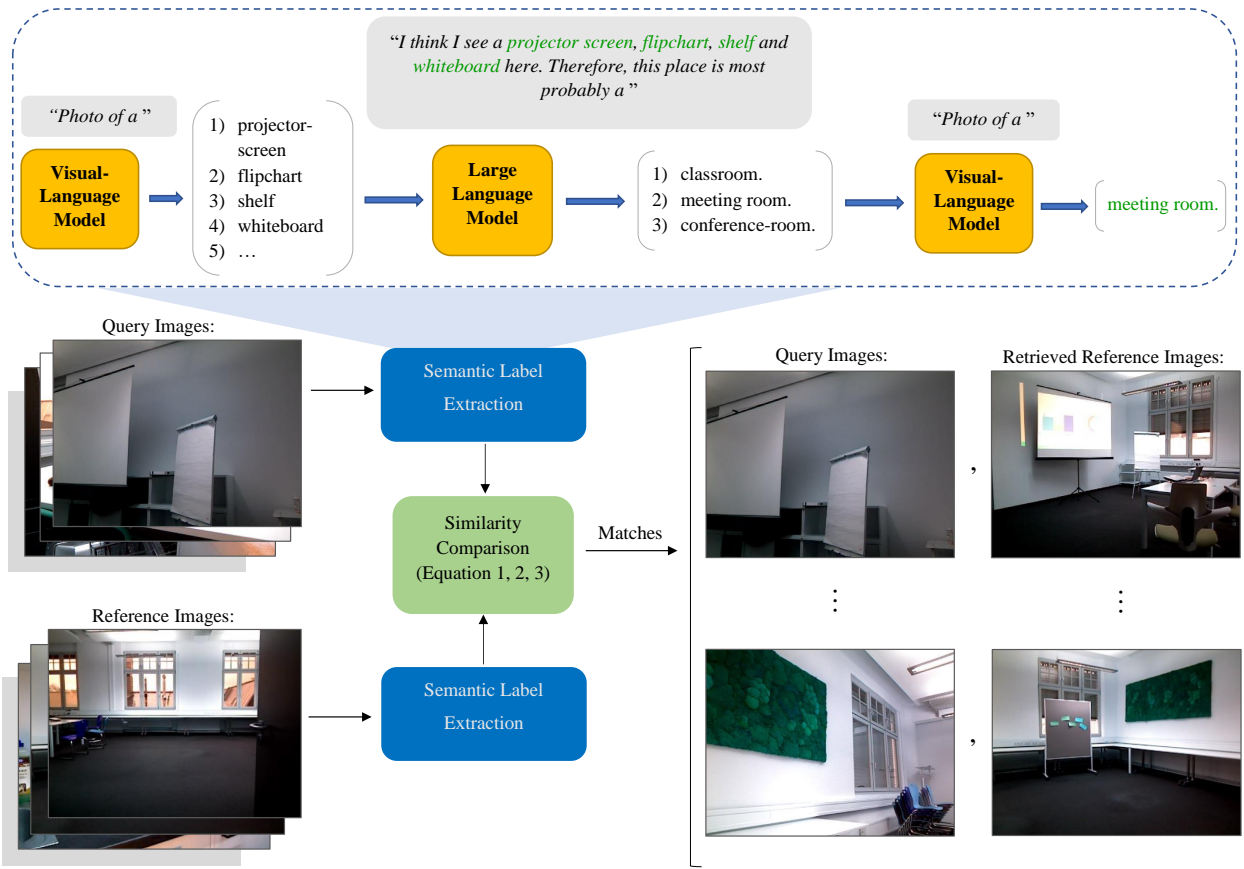}  
\caption{Our approach in a nutshell: We use a combination of Visual-Language Models (VLMs) and Large Language Models (LLMs) for vision-based localization and image retrieval under severe changes in viewpoint and scene arrangements. For each query image, we use the VLM to ground the object labels in the image and sort the objects based on their respective grounding score. We select the top-k image categories to form a prompt for the LLM to detect the room labels. To eliminate potential noise and increase the robustness of the room categorization procedure, we use the LLM to generate several room categories based on the same input prompt. We then sort these room labels by a VLM based on their grounding score. The detected objects and room labels provide the necessary semantic information for the similarity comparison between the query and each reference image. We then choose the reference frame with the highest similarity score as the retrieved image.}
\label{fig:overview}
\end{figure*}

\section{Approach}
\label{sec:approach}
In this section, we formally describe our localization approach. The  map of the robot corresponds to the sets of reference images $R$ and the corresponding camera poses $P_R$. The query image set $Q$ depicts the same location at a different time point whereby the poses $P_Q$ are unknown. Our method targets indoor environments which can be clearly separated into distinct rooms or areas, e.g., \textit{``hallway''}, \textit{``meeting room''}, or \textit{``kitchen''}. The environment can change between the recordings of $R$ and $Q$ in terms of illumination or object dynamics, and the robot can move along different trajectories. However, we assume that all rooms the robot visits in $Q$ are also present in the reference trajectory. With that, our goal is for each $I_i \in Q$ to find the best match $I_j \in R$ so that $I_i$ and $I_j$ show the same room and the pose difference between $p_i \in P_Q$ and $p_j \in P_R$ is minimal. Our approach is depicted in \autoref{fig:overview}. 


\subsection{Language-Based Descriptor}
\label{sec:language_descriptor}
We compute the semantic image descriptor in three steps: First, we utilize CLIP to detect the objects contained in the image. Based on the object labels, GPT-3 generates several proposals for room or area labels. Finally, we again use CLIP to ground the room or area labels in the image and select the one with the highest score. 

CLIP does not directly map an image to a set of object categories but performs an image-caption pairing task. It converts the image and a natural language prompt to a common embedded space where both representations are compared. For an image $I_i$ and a label $l$, $s_i(l)$ denotes the comparison score between $I_i$ and the prompt \emph{``a photo of a [$l$]''}. To detect objects with CLIP, we, therefore, need a pre-defined list $\mathcal{L}^{obj}$ of object labels. Due to the large-scale nature of the VLM, there are few restrictions on what labels can be included in that list. We take the object class labels from the MS COCO dataset~\cite{cocodataset} as a foundation that covers objects commonly found in indoor spaces. Additionally, environment-specific object categories can be suggested by the user. For each $l \in \mathcal{L}^{obj}$, we compute $s_i(l)$ and consider the five best labels as the set of detected objects $L_i^{obj}$.

In the next step, we deploy GPT-3 for classifying the scene. To that end, we plug the labels $L_i^{obj}$ into the prompt \emph{``I think I see a $[l_1,\ldots, l_5]$ here. Therefore, this place is most probably a ''} and retrieve multiple answers from the LLM. We finally select the best room label candidate $l_i^{room}$ according to the CLIP score, whereby the prompt has the same shape as for object grounding. The object labels $L_i^{obj}$ and room label $l_i^{room}$ constitute the image descriptor.

There are two reasons for the multi-step place classification described above. First, automating the room label generation by GPT-3 removes the need to pre-define the labels beforehand. Given the large training set of the LLM, this approach can provide more specific room classes (e.g., a furniture store) in addition to common ones (e.g., office, kitchen, or bedroom). This provides greater flexibility to our method and enables localization in a broad set of environments. Second, following the idea of Zheng \etal~\cite{socratic}, we use the exchange between two foundation models, trained on different datasets, to achieve a more robust room categorization.

\subsection{Similarity Comparison}
\label{sec:semantic_similarity}
We use both, the detected object categories and room labels, to compare a query image $I_i$ and the reference image $I_j$. Room labels provide a high-level yet robust estimation of the robot's location. However, most robotic applications require more accurate localization. Therefore, it is essential to use the object information in addition to the room labels to provide the algorithm with a means to localize \emph{inside} the rooms. Thus, we define the semantic similarity score $S_{i,j}^{sem}$ as the sum of object and room similarity terms:
\begin{equation}
\label{eq:sim_lang}
S_{i,j}^{\mathit{sem}} =  S_{i,j}^{\mathit{obj}} + S_{i,j}^{\mathit{\mathit{\mathit{room}}}}
\end{equation}
To calculate $S_{i,j}^{\mathit{obj}}$, we first compute the set $L_{i,j}^{\mathit{obj}}$ of object labels appearing in both images. Thus, the object similarity score is
\begin{equation}
\label{eq:sim_obj}
    S_{i,j}^{\mathit{obj}} = \sum_{l \in L_{i,j}^{\mathit{obj}}} \frac{\min(s_i(l),s_j(l))}{\max(s_i(l),s_j(l))}.
\end{equation}
For a single label $l$, the fraction term is maximized if the magnitude of $s_i(l)$ and $s_j(l)$ are similar. We choose the above formulation because the magnitude of the CLIP detection score depends on how large the object appears in the image. Therefore, $S_{i,j}^{obj}$ is higher for image pairs showing the same set of objects, where the objects have a similar distance to the camera. This helps retrieve a reference image that has a viewpoint similar to that of the query image.

The room similarity score is given by
\begin{equation}
\label{eq:sim_room}
    S_{i,j}^{\mathit{room}} = f(e(l_i^{\mathit{room}})\cdot e(l_j^{\mathit{room}})).
\end{equation}
Here, $e(l_i^{room})$ and $e(l_j^{room})$ are the normalized GPT-3 embedding vectors for the two room labels that are compared using the dot product. We do not compare the label strings directly since they are generated by the LLM and prone to noise. Often, GPT-3 finds synonymous labels for the same location, e.g., \emph{``corridor''} and \emph{``hallway''}, which, however, have a similar embedded representation. The function $f(\cdot)$ is
\begin{equation}
    f(x) = 
    \begin{cases}
    \frac{x-\theta}{1-\theta} &  x>\theta \\
    0 & \text{else}
    \end{cases}.
\end{equation}
It sets $S_{i,j}^{room}$ to zero for embeddings with a similarity below a certain threshold (we use $\theta=0.75$) and re-normalizes other values to the range $[0,1]$.

\section{Experimental Evaluation}
\label{sec:results}
In this section, we present the experimental evaluation of the proposed method. We apply our approach to real-world data that exhibits substantial environmental changes and compare the results to state-of-the-art appearance-based localization methods. The experiments are designed to demonstrate that our approach 
\begin{enumerate}
    \item can be used to improve the robustness and accuracy of vision-based localization, 
\item can easily be  extended to incorporate novel objects without the need for additional training of models, and 
\item can be utilized to identify landmarks that are relevant to the localization task.
\end{enumerate}

\begin{figure}[t]
\vspace{1ex}
\centering
\includegraphics[clip, scale=0.45]{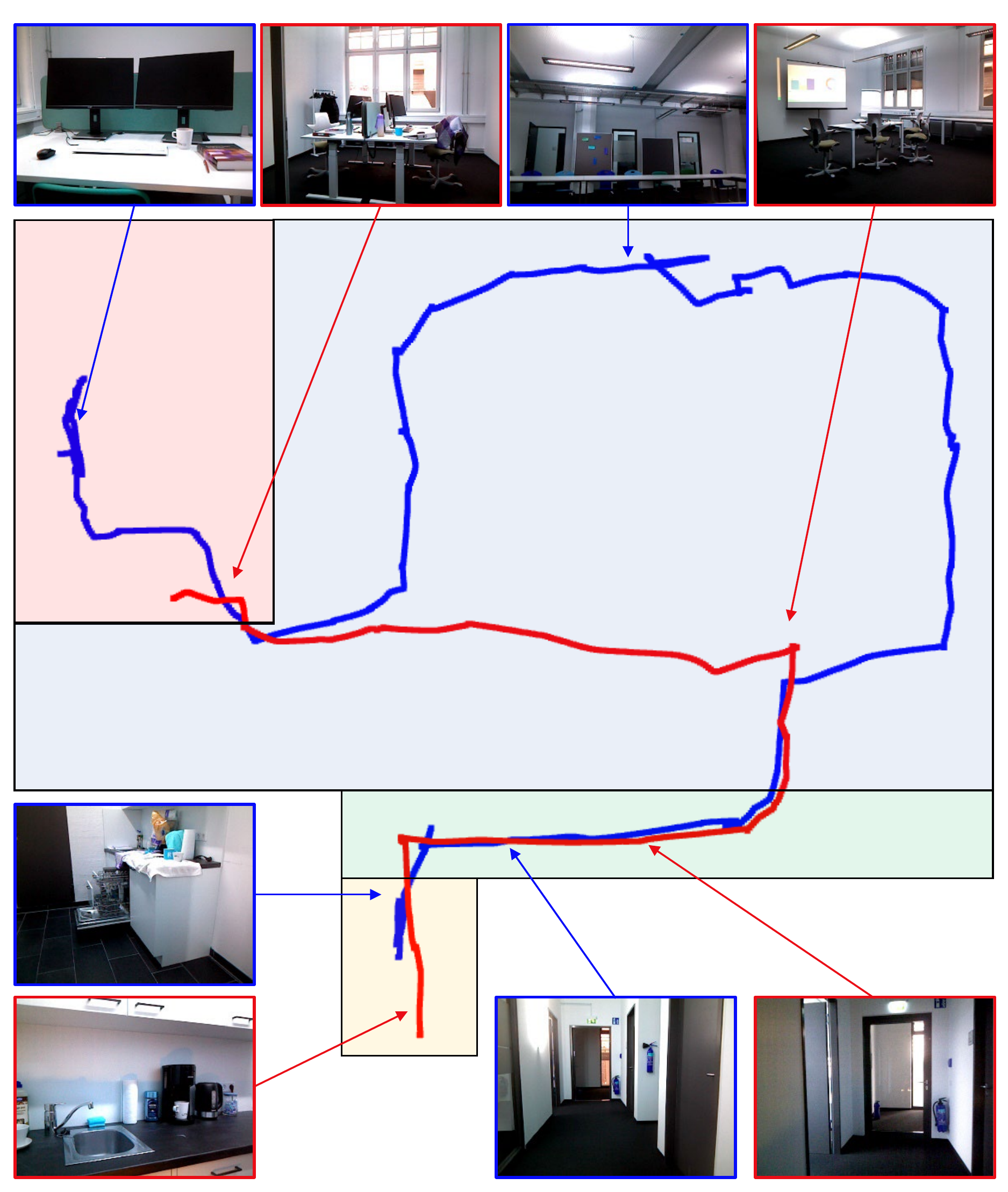}  
\caption{The reference (red) and query (blue) trajectories in dataset~1. The environment consists of four rooms, namely a kitchen (yellow), a hallway (green), a conference room (blue), and an office (red). The images depict these rooms for query and reference trajectories.}
\label{fig:ds1}
\end{figure}

\subsection{Datasets}
For evaluation, we captured two datasets each containing a reference and a query image set in two different floors of an office building. The first dataset, shown in \autoref{fig:ds1}, contains $111$ images for query and $226$ for reference while the second dataset consists of 101 images for both query and reference sets. All trajectories cover four different locations, namely a kitchen, a hallway, a large conference room, and an office. The query trajectories exhibit changes in object and furniture arrangement and have different lighting conditions. Further, the query trajectories deviate from the reference trajectories in terms of position and camera viewpoints. We used the Toyota HSR robot for recording the trajectories and the images. We aligned the two trajectories by picking image pairs from the query and reference sets that contain the same static scene from a similar view. Next, we computed the camera transformation between the images using ORB features and RANSAC. We then used the aligned trajectories for calculating the translation errors of the considered approaches. For calculating the ratio of the correct room detections, we generated the ground truth room labels by manual annotation.

\begin{figure*}[htbp]
\centering
\includegraphics[clip,trim=0cm 4cm 7cm 0cm,scale=0.65]{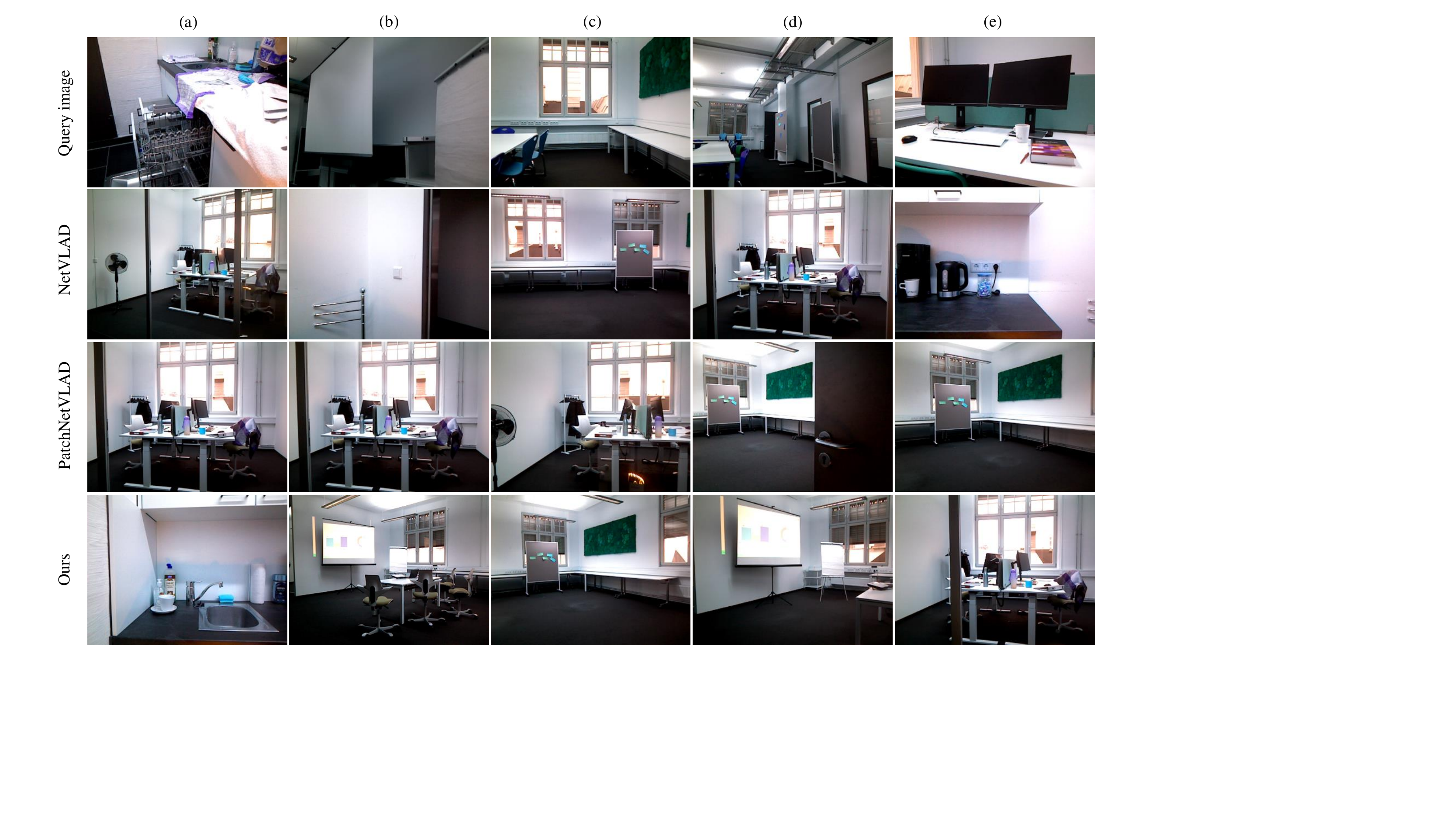}  
\caption{Comparison of query and retrieved images for different methods.}
\label{floor1_picsMatch}
\end{figure*}

\subsection{Localization}

We compare the performance of the proposed method to two state-of-the-art appearance-based localization approaches, NetVLAD~\cite{arandjelovic2016netvlad} and Patch NetVLAD~\cite{patchnetvlad}.
In our approach, we use the COCO dataset class labels as $\mathcal{L}^{obj}$ and extend it with labels of multiple objects found in our office environment, e.g., \emph{``living wall portrait''}. 

\autoref{floor1_picsMatch} shows qualitative results as a subset of the query images and their corresponding matches according to various approaches. The depicted query images emphasize different appearance variations in our dataset. Columns (a), (b), and (e) represent a combination of viewpoint changes and object re-arrangements. These appearance changes are difficult to handle for the baselines, whereas the foundation-model-based method robustly retrieves the correct references. 
The query in (c) includes an object manually added to $\mathcal{L}^{obj}$, a \emph{``living wall portrait''}. We found that this object helps our algorithm to localize within the room and to obtain a reference image with a similar viewpoint. In contrast, here NetVLAD and Patch NetVLAD focus on the distinct features of the window frame.
Lastly, the query image in (d) shows a view that is not represented in the reference trajectory. However, since the room is correctly categorized as the ``conference room'', the language-based method can successfully match this image to a reference frame in the same room. Note that the capability to easily add objects to the vocabulary is one of the strengths of our method. Given that we rely on foundation models, we generally do not need to train specific classifiers and can instead rely on the generative capabilities of the foundation models. We find this very convincing as it enables users to improve their robot's navigation performance by just providing them with a text label. 

For quantitative evaluation, we use two metrics. First, we compute the average translation error between the query image camera poses and the poses of the retrieved reference images. Second, we use the GT room labels to calculate the percentage of correct room detections. The numerical results are summarized in \autoref{1stfloor_results} and \autoref{3rdfloor_results} and are reported separately for each room and for the complete query trajectories. Our approach outperforms all baselines in the kitchen and office for both datasets. In the conference room of dataset~1, it is second to NetVLAD regarding the translation error but better on the room detection metric. In the hallway, however, our approach is behind the baselines due to the lack of distinct objects in that area. Averaged over the full trajectory of dataset~1, the proposed method has the lowest translation error with $4.98$\,meters which is $45.21\%$
lower compared to the second-best. We further achieve the highest room detection rate of $88.99\%$ on dataset~1 whereby the next best baseline achieves $54.13\%$. The results follow a similar trend for dataset~2 with an overall error of $3.96$\,meters, 
$7.26\%$ lower than the second-based, and an overall room detection rate of $84.16\%$.


\begin{table*}[h]
\caption{Comparison between language-based localization and visual feature-based methods for dataset~1}
\begin{center}
\resizebox{\textwidth}{!}{
\begin{tabular}{rrrrrrrrrrr}

\begin{tabular}{ccccccccccc}
\hline
   & \multicolumn{ 2}{c}{Kitchen} & \multicolumn{ 2}{c}{Hallway} & \multicolumn{ 2}{c}{Conference room}   & \multicolumn{ 2}{c}{Office} & \multicolumn{ 2}{c}{Total}\\
\hline
  & mean error & room detection & mean error & room detection & mean error & room detection & mean error  & room detection & mean error & room detection \\
  &  [m]& [\%]& [m] & [\%]& [m]&[\%]& [m]&[\%]& [m]&[\%]\\
\hline
NetVLAD & 4.08 & 18.18  & 0.42 & 100 & \textbf{4.19} & 22.97 & 7.73 & 37.50 & 9.09 & 39.10\\
PatchNetVLAD & 3.86 & 9.09  & \textbf{0.20} & \textbf{100} & 5.95 & 59.46 & 9.96 & 12.50 & 10.96 & 54.13\\
\algname & \textbf{2.59} & \textbf{77.27}  & 1.16 & 95.83 & 5.71 & \textbf{85.25} & \textbf{3.97} & \textbf{95.83} & \textbf{4.98} & \textbf{88.99}\\
\hline
\end{tabular}
\end{tabular}}
\end{center}

\label{1stfloor_results}
\end{table*}

\begin{table*}[h]
\caption{Comparison between language-based localization and visual feature-based methods for dataset~2}
\begin{center}
\resizebox{\textwidth}{!}{\begin{tabular}{rrrrrrrrrrr}
\label{3rdfloor_results}

\begin{tabular}{ccccccccccc}
\hline
   & \multicolumn{ 2}{c}{Kitchen} & \multicolumn{ 2}{c}{Hallway} &  \multicolumn{ 2}{c}{Conference room}  & \multicolumn{ 2}{c}{Office} &  \multicolumn{ 2}{c}{Total}\\
\hline
  & mean error & room detection & mean error & room detection & mean error & room detection & mean error & room detection  & mean error & room detection \\
   &  [m]& [\%]& [m] & [\%]& [m]&[\%]& [m]&[\%]& [m]&[\%]\\

\hline
NetVLAD & 6.49 & 47.83  & \textbf{1.10} & \textbf{100} & 7.61 & 16.67 & 4.3 & 0 & 4.27 & 59.41\\
PatchNetVLAD & 5.48 & 56.52  & 1.55 & 88.89 & 7.18 & 22.22 & 5.45 & 38.46 & 4.15 & 67.33\\
\algname & \textbf{3.19} & \textbf{91.3}  & 2.145 & 92.6 & \textbf{7.17} & \textbf{44.44} & \textbf{2.05} & \textbf{100} & \textbf{3.96} & \textbf{84.16}\\
\hline
\end{tabular}
\end{tabular}}
\end{center}
\end{table*}

\begin{table}[h]
\caption{Some of the learned landmarks for dataset~1 and their effect on reducing the mean error}
\label{landmarks1}
\begin{center}
\begin{tabular}{|l||l|}
\hline
object & error reduction [m]\\
\hline
living wall portrait & 0.71\\
\hline
projector screen & 0.66\\
\hline
conference room table & 0.51\\
\hline
computer monitor & 0.24\\
\hline
fire extinguisher & 0.19\\
\hline
exit sign & 0.16\\
\hline
\end{tabular}
\end{center}
\end{table}

\begin{table}[h]
\caption{Some of the learned landmarks for dataset~2 and their  effect on reducing the mean error}
\label{landmarks2}
\begin{center}
\begin{tabular}{|l||l|}
\hline
object & error reduction [m]\\
\hline
desk & 0.57\\
\hline
shelf & 0.47\\
\hline
exit sign & 0.44\\
\hline
projector screen & 0.17\\
\hline
dishwasher machine & 0.16\\
\hline
pinboard & 0.10\\
\hline
\end{tabular}

\end{center}
\end{table}

\subsection{Landmark Learning}
Not all detected objects are equally informative for localization, some objects are frequently displaced (e.g., cups) or are repetitive (e.g., windows). For robust localization, it is therefore beneficial to consider only static and unique landmarks. One possibility to select landmark candidates from a general object list is to prompt an LLM with the object list and the above criteria of a landmark. However, the LLM response is not grounded in the environment of the robot. Another approach is to learn the landmarks for a concrete location from data. To find the most reliable landmarks in our dataset, we repeatedly perform localization on the same data, each time eliminating one object and storing the mean translation error for that run.
We consider all objects that reduce the error by at least $0.1$\,meters as landmarks $\mathcal{L}^{lm}$. \autoref{landmarks1} and \autoref{landmarks2} show the impact of some objects on the mean error for the two different environments represented by our datasets. 

The landmark set contains the most informative objects in the considered rooms: chair, desk, and computer monitor in the office; flipchart, projector screen, desk, conference room table, and living wall portrait in the meeting room; and fire extinguisher, door, and mirror in the hallway. 
Objects frequently appearing in different rooms, e.g., dishes, increase the mean error and are therefore excluded from the landmark set. Filtering the detected object labels $L_i^{obj}$ by $\mathcal{L}^{lm}$ in the image descriptor decreases the mean localization error over the complete query trajectory to $4.71$\,meters for dataset~1 and $3.64$\,meters for dataset~2.

\subsection{Joint Semantic and Local Feature Comparison}
\label{sec:joint_score}
The language-based semantic descriptor focuses on high-level concepts like object and room classes, which makes it robust to scene dynamics and camera view variations. However, it neglects salient image gradients that are not part of an object and can therefore fail in locations with sparse object occurrences, e.g., in hallways. We thus investigate if combining the semantic matching score with the score of Patch NetVLAD improves localization. 

For each query $I_i$, we retrieve the reference $I_p$ and corresponding score $S_{i,p}^{\mathit{pat}}$ with Patch NetVLAD and then normalize these scores by dividing them by the highest score in the query trajectory, i.e., $\hat{S}_{i,p}^{\mathit{pat}} = S_{i,p}^{\mathit{pat}}/\max_j(S_{j,p}^{\mathit{pat}})$. Next, we compute a reference image $I_s$ and score $S_{i,s}^{\mathit{\mathit{sem}}}$ with the semantic approach and also the scores $\hat{S}_{i,s}^{\mathit{pat}}$ and $S_{i,p}^{\mathit{sem}}$. We choose $I_s$ as the final match if $S_{i,s}^{\mathit{sem}} + \hat{S}_{i,s}^{\mathit{pat}} > S_{i,p}^{\mathit{sem}} + \hat{S}_{i,p}^{\mathit{pat}}$ and $I_p$ otherwise. Using this joint approach improves the results on both datasets. For dataset~1 the overall mean error is reduced to $4.64$\,m and for dataset~2 it is reduced to $3.1$\,m. As expected, these improvements mainly happen in the hallway for both datasets.

\section{Conclusion and Future Work}
\label{sec:conclusion}
In this paper, we proposed \algname{}, which is a novel vision-based localization method for indoor environments based on foundation models. We utilize a visual-language model to detect objects present in an image and a large language model to categorize the depicted location. The object labels and room labels form a semantic descriptor which we use to match a query image to a set of reference images. In extensive experiments carried out in real-world environments, we demonstrate that our method is robust to severe environmental changes and variations in the camera viewpoint. As we are using foundation models, our approach does not require pre-training, training, or re-training, while still enabling the user to fine-tune it to a target environment by providing specific object labels. We furthermore presented an approach to identify the landmarks relevant to the visual localization task. 
Also, the current version of our approach requires a list of object labels that we obtained from a freely available taxonomy and that can easily be augmented with objects useful for the matching task. In future work, this could potentially be automated by a large language model that is asked to generate objects common in a given location. Through this, the robot could obtain a more robust prior for navigation without the need for user-provided landmark labels.

\bibliographystyle{IEEEtran}
\bibliography{root.bib}

\end{document}